%% file: 00_main.tex
\documentclass[runningheads]{llncs}
\usepackage[utf8]{inputenc}
\usepackage{url}
\usepackage{graphicx}
\input{cmd/pltrdy_cmd}

\begin{document}
\title{Leverage Unlabeled Data for Abstractive Speech Summarization with Self-Supervised Learning and Back-Summarization}
\titlerunning{Leverage Unlabeled Data for Abstractive Speech Summarization}
%
%
\author{Paul Tardy\inst{1, 2}
\and Louis de Seynes\inst{1}
\and François Hernandez\inst{1}
\and Vincent Nguyen\inst{1}
\and David Janiszek\inst{2,3}
\and Yannick Estève\inst{4}}
\authorrunning{P. Tardy et al}
%
\institute{Ubiqus Labs
\and LIUM -- Le Mans Université
\and Université de Paris
\and LIA -- Avignon Université
}
\maketitle              

\begin{abstract}
Supervised approaches for Neural Abstractive Summarization require large annotated corpora that are costly to build. 
We present a French meeting summarization task where reports are predicted based on the automatic transcription of the meeting audio recordings.
In order to build a corpus for this task, it is necessary to obtain the (automatic or manual) transcription of each meeting, and then to segment and align it with the corresponding manual report to produce training examples suitable for training.
On the other hand, we have access to a very large amount of unaligned data, in particular reports without corresponding transcription. Reports are professionally written and well formatted making pre-processing straightforward.
In this context, we study how to take advantage of this massive amount of unaligned data using two approaches (i) self-supervised pre-training using a target-side denoising encoder-decoder model; (ii) back-summarization i.e. reversing the summarization process by learning to predict the transcription given the report, in order to align single reports with generated transcription, and use this synthetic dataset for further training.
We report large improvements compared to the previous baseline (trained on aligned data only) for both approaches on two evaluation sets. Moreover, combining the two gives even better results, outperforming the baseline  by a large margin of $+6$ ROUGE-1 and ROUGE-L and $+5$ ROUGE-2 on two evaluation sets. 

\end{abstract}

\input{sections/10_Introduction}
\input{sections/11_relatedworks}
\input{sections/15_meetingsummarization}

\input{sections/20_selfsupervised}
\input{sections/30_backsum}

\input{sections/50_results}
\input{sections/60_discussions}
\input{sections/90_conclusion}
\bibliographystyle{splncs04}
\bibliography{bib/SPECOM2020.bib}

\end{document}

%% file: cmd/pltrdy_cmd.tex
\usepackage[dvipsnames]{xcolor}

\newcommand{\datasetname}[1]{\texttt{\small #1}}
\newcommand{\publicmeetings}{\datasetname{public\_meetings}}

\usepackage{array}
\usepackage{makecell}
\usepackage{float}

\newcommand{\subhead}[2]{\thead{\textbf{#1} \\ #2}}

\newcommand{\best}[1]{\textbf{#1}}

\usepackage{booktabs}


%% file: sections/10_Introduction.tex
\section{Introduction}

Automatic Meeting Summarization is the task of writing the report corresponding to a meeting. 
We focus on so-called \textit{exhaustive reports}, which capture, in a written form, all the information of a meeting, keeping chronological order and speakers' interventions. Such reports are typically written by professionals based on their notes and the recording of the meeting.

Learning such a task with a supervised approach requires building a corpus, more specifically (i) gathering $(transcription, report)$ pairs; (ii) cutting them down into smaller segments; (iii) aligning the segments; and, finally, (iv) training a model to predict the report from a given transcription.

We built such a corpus using Automatic Speech Recognition (ASR) to generate transcription, then aligning segments manually and automatically \cite{Tardy2020}. This process is costly to scale, therefore, a large majority of our data remains unaligned. In particular, we have access to datasets with very different orders of magnitude; we have vastly more reports ($10^6$) than aligned segments ($10^4$). Reports are supposedly error-free and well formatted, making it easy to process on larger scale.

In this work, we focus on how to take advantage of those unaligned data. We explore two approaches, first a self-supervised pre-training step, which learns internal representation based on reports only, following BART \cite{Lewis2019a}; then we introduce meeting \textit{back-summarization} which is to predict the transcription given the report, in order to generate a synthetic corpus based on unaligned reports; finally, we combine both approaches.

%% file: sections/11_relatedworks.tex
\section{Related Work}


Abstractive Summarization has seen great improvements over the last few years following the rise of neural sequence-to-sequence models~\cite{Cho2014,Sutskever2014}. Initially, the field mostly revolved around headline generation \cite{Rush2015}, then multi-sentence summarization on news datasets (e.g. CNN/DailyMail corpus~\cite{Nallapati2016}). Further improvements include pointer generator \cite{Nallapati2016,see2017} which learns whether to generate words or to copy them from the source; attention over time \cite{Nallapati2016,see2017,Paulus2017}; and hybrid learning objectives \cite{Paulus2017}. Also, \textit{Transformer} architecture~\cite{Vaswani2017} has been used for summarization~\cite{Gehrmann2018,ziegler2019}. In particular, our baseline architecture, hyper-parameters and optimization scheme are similar to~\cite{ziegler2019}.
All these works are supervised, thus, useful for our baseline models on aligned data.

In order to improve performances without using aligned data, one approach is self-supervision. 
Recent years have seen a sudden increase of interest about self-supervised large Language Models (LM)~\cite{devlin-etal-2019-bert,Radford2019} since it showed impressive transfer capabilities. This work aims at learning general representations such that they can be fine-tuned into more specific tasks like Question Answering, Translation and Summarization. Similarly, BART~\cite{Lewis2019a} proposes a pre-training based on input denoising i.e. reconstructing the correct input based on a noisy version. BART is no more than a standard \textit{Transformer} model with noise functions applied to its input, making it straightforward to implement, train and fine-tune.

Another way of using unaligned data is to synthetically align it. While this is not a common practice for summarization (at the best of our knowledge) this has been studied for Neural Machine Translation as Back-Translation~\cite{Sennrich2016_backtranslation,Poncelas2018,Edunov2018}. We propose to apply a similar approach to summarization to generate synthetic transcriptions for unlabeled reports.


%% file: sections/15_meetingsummarization.tex
\section{Meeting summarization}


\subsection{Corpus Creation}
Using our internal Automatic Speech Recognition (ASR) pipeline, we generate transcriptions from meeting audio recordings. 

Our ASR model has been trained using the Kaldi toolkit~\cite{Povey2011-kaldi} and a data set of around $700$ hours of meeting audio. Training data was made of \textit{(utterance text, audio sample)} pairs that were aligned with a Kaldi functionality to split long audio files into short text utterances with their corresponding audio parts (section 2.1 of \cite{Hernandez2018}). We end up with a train set of approximately 476k pairs. We used a chain model with a TDNN-F~\cite{Povey2018-tdnn} architecture, $13$ layers of size $1024$, an overall context of ($-28,28$) and around $1.7$ million parameters. We have trained our model for 20 epochs with batch-normalization and L2-regularization. On the top of that, we added an RNNLM~\cite{Sak2014} step to rescore lattices from the acoustic model. This model consisted of three TDNN layers, two interspersed LTSMP layers resulting in a total of around 10 million parameters.

Our raw data, consisting in pairs of audio recordings (up to several hours long) and exhaustive reports (dozens of pages) requires segmentation without loss of alignment, i.e. producing smaller segments of transcription and report while ensuring that they match. This annotation phase was conducted by human annotators using automatic alignment as described in \cite{Tardy2020}. We refer to these aligned corpuses as, respectively, \textit{manual} and \textit{automatic}.

 While this process allowed us to rapidly increase the size of the dataset, it is still hard to reach higher orders of magnitude because of time and the resource requirements of finding and mapping audio recordings with reports, running the Automatic Speech Recognition pipeline (ASR) and finally the automatic alignment process.
 
 \subsection{Baseline \& Setup}
 Since automatic alignment has been showed to be beneficial~\cite{Tardy2020}, we train baselines on both manually and automatically aligned data. Models are standard \textit{Transformer}~\cite{Vaswani2017} of $6$ layers for both encoder and decoder, $8$ attention heads and $512$ dimensions for word-embeddings. Hyper-parameters and optimization are similar to the baseline model of \cite{ziegler2019} (just switching-off copy-mechanism and weight sharing as discussed in \ref{sec:reducing_extractivity}).
 During pre-processing, text is tokenized using \textit{Stanford Tokenizer}~\cite{manning2014}, \textit{Byte Pair Encoding} (BPE)~\cite{sennrich2016-subwords-bpe}. Training and inference uses \textit{OpenNMT-py}\footnote{\url{https://github.com/OpenNMT/OpenNMT-py}}~\cite{2017opennmt}.

In this paper, we study how simple models -- all using the same architecture -- can take advantage of the large amount of single reports (i.e. not aligned with transcription).


%% file: sections/20_selfsupervised.tex
\section{Self-supervised Learning}
Self-supervised learning aims to learn only using unlabeled data. It has recently been applied -- with great success -- to Natural Language Processing (NLP) as a pre-training step to learn word embeddings using \textit{word2vec}\cite{Mikolov2013-word2vec}, encoder representations with \textit{BERT}\cite{devlin-etal-2019-bert} or an auto-regressive decoder with \textit{GPT-2} \cite{Radford2019}.

More recently, BART \cite{Lewis2019a} proposed to pre-train a full \textit{Transformer} based encoder-decoder model with a denoising objective. They experimented with various noise functions and showed very good transfer performances across text generation tasks like question answering or several summarization datasets. When applied to summarization, the best results were obtained with \textit{text-infilling} (replacements of continuous spans of texts with special mask tokens) and \textit{sentence permutation}. 

The model is pre-trained on \textit{reports} only, thus not requiring alignment. It allows us to use our unlabeled reports at this stage. We use the same setup as BART, i.e. two noise functions: (i) \textit{text-infilling} with $p=0.3$ and $\lambda=3$; and (ii) \textit{sentence permutation} with $p=1$.

Since the whole encoder-decoder model is pre-trained, the fine-tuning process is straightforward: we just reset the optimizer state and train the model on the \textit{transcription} to \textit{report} summarization task using aligned data (\textit{manual+auto}).


%% file: sections/30_backsum.tex
\section{Back-summarization}
\label{sec:backsummarization}
Instead of considering single documents as unlabeled data, we propose to reverse the summarization process in order to generate synthetic transcriptions for each report segment. We call this process \textit{back-summarization} in reference to the now well known \textit{back-translation} data augmentation technique for Neural Machine Translation~\cite{Sennrich2016_backtranslation}. 

\textit{Back-summarization} follows three steps:
\begin{enumerate}
    \item \textbf{backward training}: using manually and automatically aligned datasets ($man+auto$), we train a back-summarization model that predicts the source (= \textit{transcription}) given the target (= \textit{report}).

    \item \textbf{transcription synthesis}: synthetic sources (=\textit{transcriptions}) are generated for each single target with the backward model, making it a new aligned dataset, denoted \textit{back}.

    \item \textbf{forward training}: using all three datasets ($man+auto+back$), we train a summarization model.
\end{enumerate}

Backward and forward models are similar to the baseline in terms of architecture and hyper-parameters. Synthetic transcriptions are generated by running inference on the best backward model with respect to the validation set in terms of ROUGE score while avoiding models that copy too much from the source (more details in \ref{sec:reducing_extractivity}). Inference uses beam search size $5$ and trigram repetition blocking~\cite{Paulus2017}. 

During forward training, datasets have different weights. These weights set how the training process iterates over each dataset. For example, if weights are $(1, 10, 100)$ respectively for $(man, auto, back)$, it means that for each \textit{man} example, the model will also train over $10$ \textit{auto} and $100$ \textit{back} examples. This is motivated by \cite{Edunov2018} that suggests (in section 5.5) \textit{upsampling} -- i.e. using more frequently -- aligned data in comparison to synthetic data. To be comparable to other models, we set \textit{man} and \textit{auto} weights close to their actual size ratio i.e. $(2, 7)$. For synthetic data we experiment with $100$, giving an approximate \textit{upsample} rate of $6$.

%% file: sections/50_results.tex
\section{Results}
Experiments are conducted against two reference sets, the \publicmeetings~ test set and the valid set.
Summarization models are evaluated with ROUGE~\cite{Lin2004}. We also measure what proportion of the predictions is copied -- denoted by copy\% -- from the source based on the ROUGE measure between the source and the prediction. Datasets for both training and evaluation are presented in table~\ref{table:datasets_ubiqus}.

\subsection{Reducing extractivity bias}
\label{sec:reducing_extractivity}
While training and validating models for back-summarization we faced the extractivity bias problem. It is a common observation that even so-called abstractive neural summarization models tend to be too extractive by nature, i.e. they copy too much from the source. This is true for first encoder-decoder models \cite{see2017} but also for more recent -- and more abstractive -- models like BART~\cite{Lewis2019a}.

Back-summarization amplifies this bias since it relies on two summarization steps: backward, and forward, both facing the same problem. At first, we trained typical \textit{Transformer} models for summarization that involve copy-mechanism \cite{see2017} and weight sharing between (i) encoder and decoder embeddings, (ii) decoder embedding and generator, similar to the baseline of \cite{ziegler2019}. When evaluating against the validation set -- reference copy\%: $55.38$ --, the predictions had too much copy: $62.64\%$, ($+7.26$).

We found that turning off both copy mechanism and weight sharing reduces the copy\% to $54.65$, ($-0.73$). Following these interesting results, we also apply this to other models.



\subsection{Datasets}

Datasets used in this paper are presented in table \ref{table:datasets_ubiqus}. We refer to the human annotated dataset as \textit{manual} and the automatically aligned data as \textit{automatic}. \textit{Backsum} data consists of single reports aligned with synthetic transcription using \textit{back-summarization} (see section \ref{sec:backsummarization}). 
The validation set -- \textit{valid} -- is made of manually aligned pairs excluded from training sets.
Finally, we use \publicmeetings~\cite{Tardy2020} as a test set. We observe differences between \publicmeetings\, in term of lengths of source/target and extractivity. We hypothesize that constraints specific to this corpus -- i.e. the use of meetings with publicly shareable data only  --  introduced bias. For this reason, we conduct evaluation on both \publicmeetings\, and the validation set.


    \begin{table}
    \begin{center}
    \caption{About Ubiqus Datasets: number of examples, lengths of source and target (words in average, first decile $d_1$ and last $d_9$), and the extractivity measured with ROUGE between target and source. (*) Extractivity of the back-summarization dataset is measured on a subset of $10,000$ randomly sampled examples }
    \label{table:datasets_ubiqus}
    \setlength{\tabcolsep}{6pt}
    \begin{tabular}{l*4{c }} 
    
    \toprule
        \subhead{Dataset}{\hfill}
            & \subhead{\#Pairs}{\hfill}
            & \subhead{src}{($avg, [d_1, d_9])$}
            & \subhead{tgt}{($avg, [d_1, d_9])$}
            & \subhead{Copy \%}{$R1$}
    \\\midrule  Manual 
                    & 21k
                    & 172, [42, 381]
                    & 129, [25, 297]
                    & 55.45
    \\          Automatic 
                    & 68k
                    & 188, [45, 421]
                    & 130, [25, 302]
                    & 54.72
    \\          Valid
                    & 1k
                    & 178, [40, 409]
                    & 144, [22, 294]
                    & 55.38
    \\          Back
                    & 6.3m
                    & 232, [53, 509]
                    & 90, [43, 153]
                    & 63.09*
    \\          \publicmeetings
                    & 1060
                    & 261, [52, 599]
                    & 198, [33, 470]
                    & 75.84
    \\\bottomrule
    \end{tabular}
    \end{center}
    \end{table}

\subsection{Summarization}
Summarization results are presented in table \ref{table:results_summarization_publicmeetings} on the test set and table \ref{table:results_summarization_valid} on the valid set.

Self-supervised pre-training outperforms the baseline by a large margin: $+2.7$ ROUGE-1 and ROUGE-2 on the test set and up to $+4$ on the validation set, for both ROUGE-1 and ROUGE-2.

\textit{Back-summarization} models -- without pre-training -- also improve the baseline, by a larger margin: $+2.98$ R1, $+2.39$ R2 on the test set, and up to $+5.34$ R1 and $+4.80$ R2 on the validation set.

Interestingly, mixing both approaches consistently outperforms other models, with large improvement on both the test set ($+5.71$ R1 and $+4.64$ R2) and the validation set ($+6.40$ R1 and $+5.16$ R2).

In addition to the ROUGE score, we also pay attention to how extractive the models are. As discussed in section \ref{sec:reducing_extractivity}, we notice that the models are biased towards extractive summarization since they generate predictions with higher copy\% on both reference sets (respectively $75.85$ copy\% and  $55.38$ copy\% for test and valid).
Even though we paid attention to this extractivity bias during the \textit{back-summarization} evaluation, we still find \textit{backsum} models to increase copy\% more than self-supervised models.

    \begin{table}
    \begin{center}
    \caption{Scores on the \publicmeetings~test set}
    \label{table:results_summarization_publicmeetings}
    \setlength{\tabcolsep}{6pt}
    \begin{tabular}{lll*2{c}} 
    \toprule
        \subhead{Model}{\hfill}
            & \subhead{Training Steps}{\hfill}
            & \subhead{ROUGE Score (F)}{$(R1, R2, RL)$}
            & \subhead{Copy \%}{$R1$}
    \\\midrule    Baseline
                    & 8k
                    & 52.31 / 34.00 / 49.70    
                    & 79.36
    \\  SelfSup
                    & 4k (pre-trained 50k)
                    & 55.08 / 36.76 / 52.43    
                    & 83.72
    \\  Backsum
                    & 6k
                    & 55.29 / 36.39 / 52.89    
                    & 89.24
    \\  Both
                    & 6k (pre-trained 50k)
                    & \best{58.02} / \best{38.64} / \best{55.56}    
                    & 90.77
    \\\bottomrule
    \end{tabular}
     \end{center}
    \end{table}
    \begin{table}
    \begin{center}
    \caption{Scores on the validation set}
    \label{table:results_summarization_valid}
    \setlength{\tabcolsep}{6pt}
    \begin{tabular}{ll*3{c}} 
    \toprule
        \subhead{Model}{\hfill}
            & \subhead{Training Steps}{\hfill}
            & \subhead{ROUGE Score (F)}{$(R1, R2, RL)$}
            & \subhead{Copy. \%}{$R1$}
    \\\midrule    Baseline
                    & 6k
                    & 33.83 / 15.86 / 31.05    
                    & 74.25
    \\  SelfSup
                    & 4k (pre-trained 50k)
                    & 37.94 / 19.16 / 34.86
                    & 78.61
    \\  Backsum
                    & 10k
                    & 39.17 / 20.06 / 36.19    
                    & 86.96
    \\  Both
                    & 6k (pre-trained 50k)
                    & \best{40.23} / \best{21.02} / \best{37.26}    
                    & 88.03
    \\\bottomrule
    \end{tabular}
     \end{center}
    \end{table}

%% file: sections/60_discussions.tex
\section{Discussion}

\subsubsection{Copy\% on different datasets}
Running evaluations on two datasets makes it possible to compare how models behave when confronted with different kinds of examples. In particular, there is a 20\% difference in terms of copy rate between evaluation sets (\publicmeetings\, and valid). On the other hand, for any given model, there is a difference of less than $5$ copy\%. 

This suggests that models have a hard time adapting in order to generate more abstractive predictions when they should. In other words, models are bad at predicting the expected copy\%.

\subsubsection{ROUGE and copy\%.}
Our results suggest that better models, in terms of the ROUGE score also have a higher copy\%. This poses the question of which metric we really want to optimize. On the one hand, we want to maximize ROUGE, on the other  we want our prediction to be abstractive i.e. stick to the reference copy rate. These two objectives currently seem to be in opposition, making it hard to choose the most relevant model. For example, on the test set (table \ref{table:results_summarization_publicmeetings}) Backsum performs similarly to SelfSup with respect to ROUGE but has a much higher copy\%, which would probably be penalized by human evaluators.


%% file: sections/90_conclusion.tex
\section{Conclusion}
We presented a French meeting summarization task consisting in predicting exhaustive reports from meeting transcriptions.
In order to avoid constraints from data annotation, we explore two approaches that take advantage of the large amount of available unaligned reports: (i) self-supervised pre-training on reports only; (ii) back-summarization to generate synthetic transcription for unaligned reports. 
Both approaches exhibit encouraging results, clearly outperforming the previous baseline trained on aligned data. Interestingly, combining the two techniques leads to an even greater improvement compared the baseline :  $+6$/$+5$/$+6$ ROUGE-1/2/L